\documentclass{article}

\PassOptionsToPackage{numbers, compress}{natbib}

\usepackage[preprint]{neurips_2021}

\usepackage{multirow}
\usepackage{graphicx}
\usepackage{lscape}
\usepackage[utf8]{inputenc} 
\usepackage[T1]{fontenc}    
\usepackage{hyperref}       
\usepackage{url}            
\usepackage{booktabs}       
\usepackage{amsfonts}       
\usepackage{nicefrac}       
\usepackage{microtype}      
\usepackage{xcolor}         

\title{Introducing a Family of Synthetic Datasets for Research on Bias in Machine Learning}

\author{
  William Blanzeisky\\
  School of Computer Science\\
  University College Dublin\\
  \texttt{william.blanzeisky@ucdconnect.ie} \\
   \And
  P\'{a}draig Cunningham \\
  School of Computer Science\\
  University College Dublin\\
  \texttt{padraig.cunningham@ucd.ie} \\
    \AND
 Kenneth Kennedy\\
 Credit Scoring Consultant\\
 \texttt{kennedykenneth@gmail.com}
 }

\begin{document}

\maketitle

\begin{abstract}
A significant impediment to progress in research on bias in machine learning (ML) is the availability of relevant datasets. This situation is unlikely to change much given the sensitivity of such data. For this reason there is a role for synthetic data in this research. In this short paper we present one such family of synthetic data sets. We provide an overview of the data, describe how the level of bias can be varied and present a simple example of an experiment on the data.   
\end{abstract}

\section{Introduction}
Bias and fairness in ML has received a lot of research attention in recent years. This is due at least in part to a number of high-profile issues with deployed ML systems \cite{holstein2019improving}. One of the challenges for research in this area is the availability of appropriate datasets. In this paper we present a family of synthetic datasets that we hope will address this issue.

Bias can occur in ML classification when we have a desirable outcome  $Y = 1$ and a sensitive feature $S$ where $S = 1$ is the majority class and $S = 0$ is the sensitive minority. 
Bias may be quantified in various ways \cite{caton2020fairness}: one of the accepted measures is Disparate Impact (DIs) \cite{feldman2015certifying}:
\begin{equation}\label{eqn:DI}
   \mathrm{DI}_S \leftarrow \frac{P[\hat{Y}= 1 | S = 0]}{P[\hat{Y} = 1 \vert S = 1]} < \tau 
\end{equation}
that is, the probability of good outcomes $\hat{Y}$ predicted for the sensitive minority are less than those for the majority. A $\tau$ threshold of 0.8 would represent the 80\% rule: scenarios with $\tau < 0.8$ would be considered clearly unfair. 

It is often the case the disparate impact arises due to shortcomings in the training data representing discriminatory practice in the past. It may also happen that the ML algorithm itself is biased failing to pick up patterns in the data due to model \emph{underfitting}. This is termed underestimation \cite{Cunningham_2021}. An underestimation score ($\mathrm{US}_{S}$) in line with $\mathrm{DI}_S$ would be:
\begin{equation}\label{eqn:US_S1}
    \mathrm{US}_{S} \leftarrow \frac{P[\hat{Y}= 1 | S = 0]}{P[Y = 1 | S = 0]}
\end{equation}
A poor underestimation score ${US}_{S} < 1$ would indicate that the predicted outcomes $\hat{Y}$ for the minority are out of line with the actual outcomes $Y$ in test data. 

Even though there are many datasets available for ML research there are not many datasets suitable for testing impacts and remedies for disparate impact or underestimation. Feldman \emph{et al.} \cite{feldman2015certifying} introduced a synthetic dataset in 2015 to help address this issue. 
While this dataset has been used in a wide range of studies on bias in ML it is limited in that it is a fairly simple dataset. The outcome is whether a student will be accepted for a college course; there are just three inputs,  SAT score, IQ and gender. By contrast the synthetic dataset we present here has 18 features with complex interactions between them \cite{kennedy2011framework}. This dataset is presented in the next section and some basic experiments are presented in \ref{sec:experiments}.

\section{The Data}\label{sec:TheData}

In this section, we present an overview of the synthetic credit scoring dataset summarised in Table \ref{tab:my-table}. Each sample in the dataset represents a residential mortgage applicant's information and the class label is generated based on a threshold to distinguish between those likely to repay and those likely to default on their financial obligation. Hence, the classification objective is to assign borrowers to one of two groups: good or bad. The \textit{framework}\footnote{\url{https://www.researchgate.net/publication/241802100_A_Framework_for_Generating_Data_to_Simulate_Application_Scoring}} used to generate these synthetic datasets is explained in \cite{kennedy2011framework}. 


In line with other research on bias in ML, we select \textit{Age} as  sensitive attribute. Given that the context is mortgage applications, we categorize younger people as the underprivileged group $S = 0$ and older people as the privileged group $S = 1$. Since Age is a continuous variable, we can vary the magnitude of under-representation by adjusting the threshold of separating privileged $S = 1$ and unprivileged group $S = 0$. In addition, we can also vary the level of class imbalance in the dataset by adjusting the threshold on the class labels.

\subsection{Generating Specific Datasets}
We split the dataset generated in Section \ref{sec:TheData} into three subsets: small, medium and large base datasets, which contain 3,493, 13,231 and 37,607 samples respectively. Through user-defined parameters, it is possible to vary the level of bias by adjusting the threshold on the class label $Y$ to distinguish between those likely to repay $Y = 1$ and those likely to default on their financial obligation $Y = 0$ (class imbalance), and on the sensitive feature \textit{Age} to separate privileged $S = 1$ and unprivileged group $S = 0$ (feature imbalance). In addition, a Gaussian random noise can also be added to \textit{Score} to increase classification complexity: this is controlled by a user-defined parameter. To ensure reproducibility, we have provided access to all of the data and code used in this article at the author’s \textit{Github}\footnote{\url{https://github.com/williamblanzeisky/SBDG}} page.

\subsection{The Exemplar Dataset}
We also provide exemplar datasets where  specific thresholds have been applied. The thresholds have been set so that there is 25\% class imbalance and 30\% feature imbalance. In the next section, we will conduct a set of experiments on one of these exemplar datasets to show clear evidence of negative legacy and underestimation. 

\section{Sample Experiments}\label{sec:experiments}

\begin{figure}[h]
     \centering
     \includegraphics[width=0.7\linewidth]{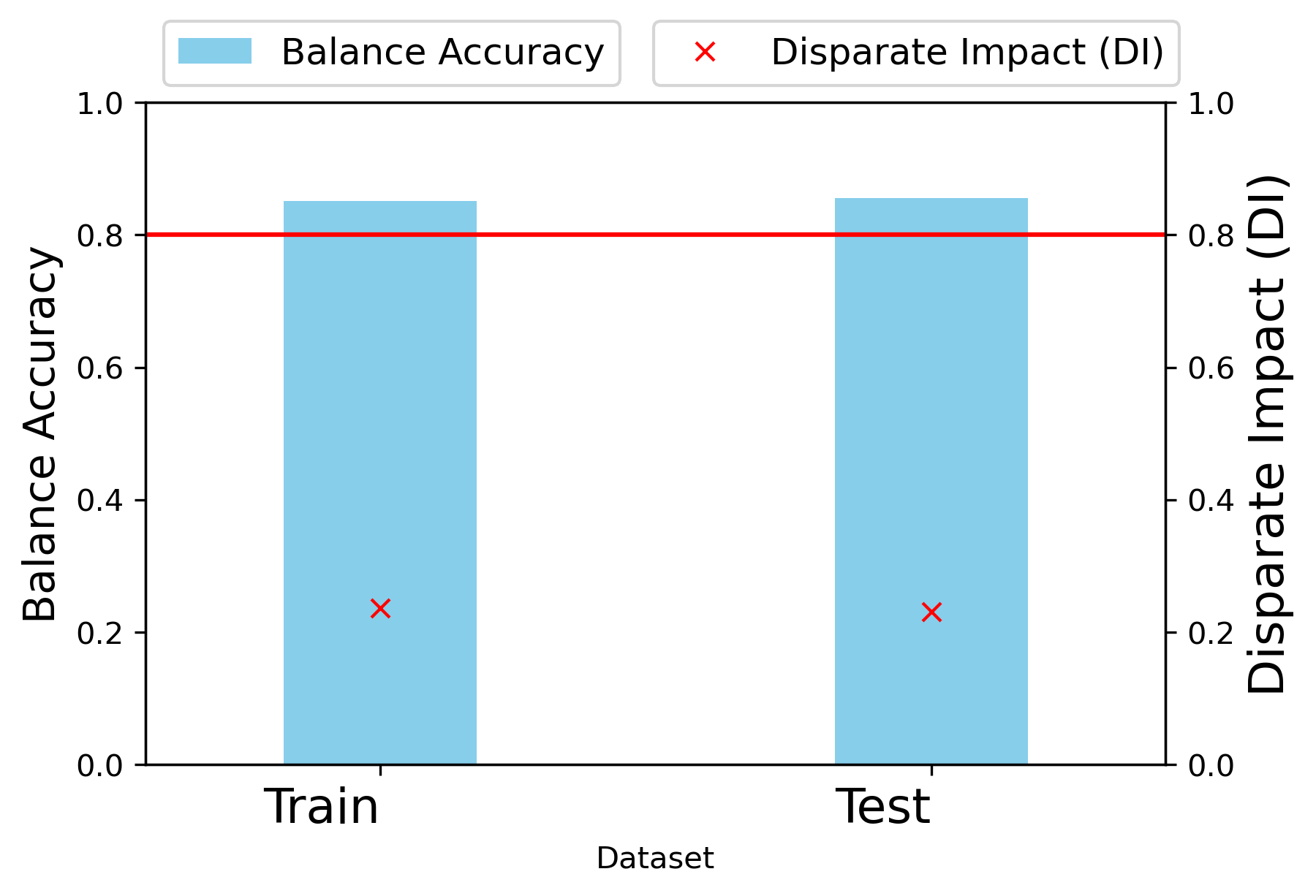}
     \caption{A demonstration of negative legacy on the train and test set. The red horizontal line represents 80\% rule of Disparate Impact.}
     \label{fig:negativelegacy}
\end{figure}

Using the scikit-learn logistic regression classifier, we show a clear evidence of negative legacy on the medium exemplar dataset. 70\% of the observations in the dataset are used for training, and the remaining 30\% are reserved for model testing. Figure \ref{fig:negativelegacy} illustrates the disparate impact score $DI_{S}$ and balance accuracy for the train and test set. We see that the logistic regression model does not satisfy the 80\% rule for disparate impact.

\begin{figure}[h]
     \centering
     \includegraphics[width=0.7\linewidth]{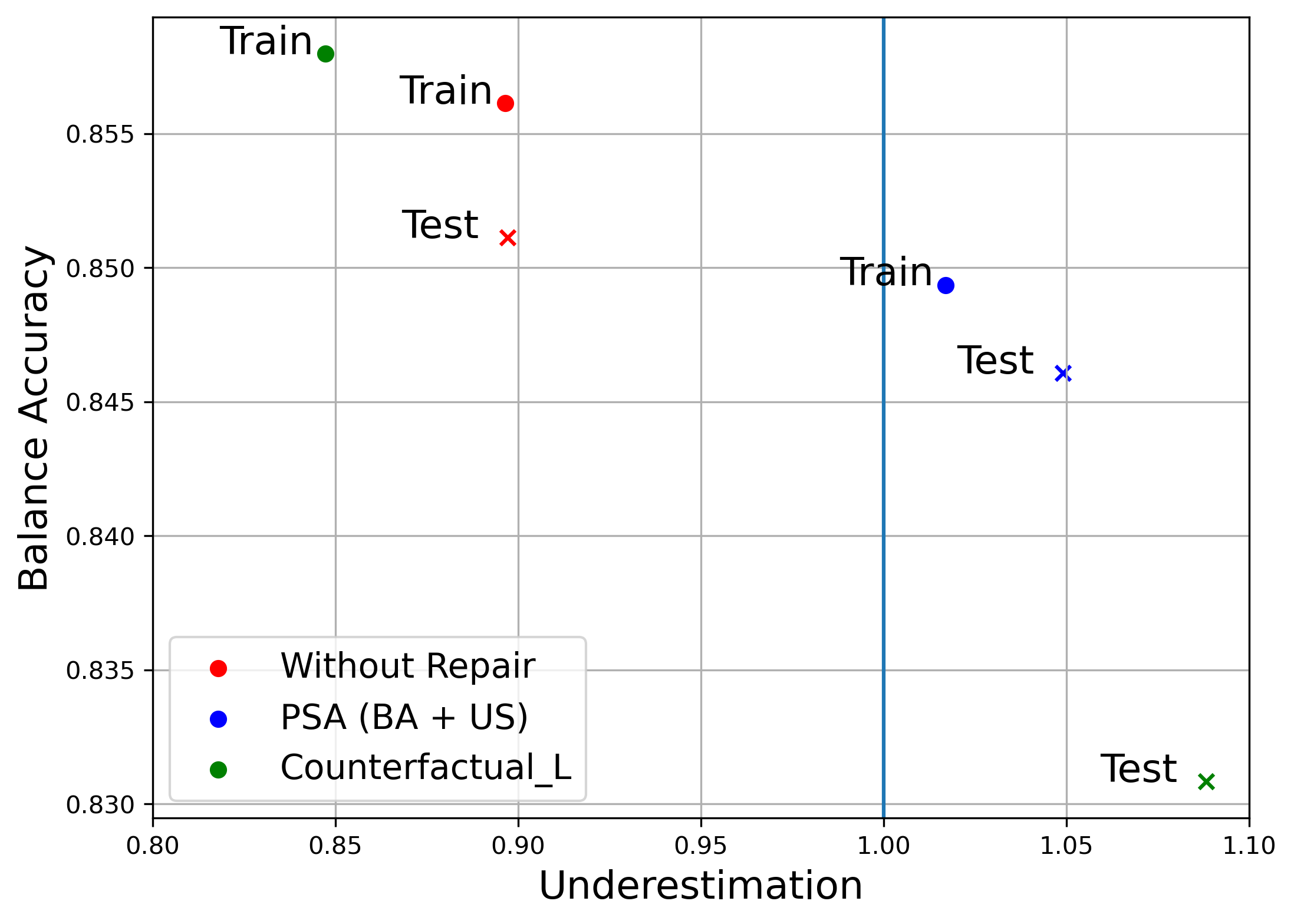}
     \caption{Evaluation of two remediation strategies to fix underestimation. It is clear that both strategies (PSA (BA + US) and Counterfactual$_L$) fixes underestimation without much loss in balance accuracy. The blue vertical line $US_S = 1.0$ represents the desired underestimation score.}
     \label{fig:eval}
\end{figure}

In addition, we conducted some experiments to demonstrate the effectiveness of current remediation strategies in the literature to fix underestimation. Specifically, we evaluate two categories of mitigation techniques: by adding counterfactuals (pre-processing) \cite{DBLP:journals/corr/abs-2104-14014} and using Pareto Simulated Annealing to simultaneously optimize a classifier on balance accuracy and underestimation (in-processing) \cite{blanzeisky2021using}. Figure \ref{fig:eval} shows the effectiveness of these strategies on the medium exemplar dataset. We can see that the in-processing strategy (PSA (BA+US)) outperforms the other strategies in terms of underestimation while maintaining comparable balance accuracy.

\begin{landscape}
\begin{table}[]
\begin{tabular}{|l|llll}
\hline
\multicolumn{1}{|c|}{\textbf{Type}} &
  \multicolumn{1}{c|}{\textbf{Variable}} &
  \multicolumn{1}{c|}{\textbf{Description}} &
  \multicolumn{1}{c|}{\textbf{Range/Values}} &
  \multicolumn{1}{c|}{\textbf{Label}} \\ \hline
\multicolumn{1}{|c|}{\multirow{7}{*}{Categorical}} &
  \multicolumn{1}{l|}{Location} &
  \multicolumn{1}{l|}{Location of purchased dwelling} &
  \multicolumn{1}{l|}{\begin{tabular}[c]{@{}l@{}}(1) Dublin; (2) Cork; (3) Galway; (4) Limerick;\\ (5) Waterford; (6) Other\end{tabular}} &
  \multicolumn{1}{l|}{Location} \\ \cline{2-5} 
\multicolumn{1}{|c|}{} &
  \multicolumn{1}{l|}{Employment Sector} &
  \multicolumn{1}{l|}{Borrower's employment sector} &
  \multicolumn{1}{l|}{\begin{tabular}[c]{@{}l@{}}(1) Agriculture, Forestry and fishing; (2) Construction;\\  (3) Wholesale and Retail; (4) Transportation and Storage; \\ (5) Hospitality; (6) Information and Communication; \\ (7) Professional, Scientific and Technical; (8) Admin and \\ Support services; (9) Public Admin.; (10) Education; \\ (11) Health; (12) Industry; (13) Financial; (14) Other\end{tabular}} &
  \multicolumn{1}{l|}{EmpSector} \\ \cline{2-5} 
\multicolumn{1}{|c|}{} &
  \multicolumn{1}{l|}{Occupation} &
  \multicolumn{1}{l|}{Employment activity of the borrower} &
  \multicolumn{1}{l|}{\begin{tabular}[c]{@{}l@{}}(1) Manager, Administrator, and Professional (henceforth MAP); \\ (2) Associate professional and technical, Clerical and secretarial, \\ Personal and protective service, and Sales (henceforth Office); \\ (3) Craft and related (henceforth Trade); (henceforth Office); \\ (4) Craft and related (henceforth Trade); (5) Other manual\\ operators (henceforth Farmer); (6) Self Employed\end{tabular}} &
  \multicolumn{1}{l|}{Occ} \\ \cline{2-5} 
\multicolumn{1}{|c|}{} &
  \multicolumn{1}{l|}{Household} &
  \multicolumn{1}{l|}{Family Composition} &
  \multicolumn{1}{l|}{\begin{tabular}[c]{@{}l@{}}(1) 1 Adult, no child \textless 18; (2) 1 Adult, 1+ child \textless 18; \\ (3) 2 Adults, no child \textless 18; (4) 3+ adults, no child \textless 18; \\ (5) 2 Adults, 1+ child \textless 18; (6) Other\end{tabular}} &
  \multicolumn{1}{l|}{HouseComp} \\ \cline{2-5} 
\multicolumn{1}{|c|}{} &
  \multicolumn{1}{l|}{Education} &
  \multicolumn{1}{l|}{Highest level of formal education} &
  \multicolumn{1}{l|}{\begin{tabular}[c]{@{}l@{}}(1) Primary or below (PB); (2) Lower secondary (LS); (3) Higher \\ secondary (HS); (4) Post leaving certificate (PLC); (5) Third level \\ non-honours degree (TLND); (6) Third level honours degree or \\ above (TLHD); (7) Other\end{tabular}} &
  \multicolumn{1}{l|}{Education} \\ \cline{2-5} 
\multicolumn{1}{|c|}{} &
  \multicolumn{1}{l|}{Label} &
  \multicolumn{1}{l|}{Risk of defaulting} &
  \multicolumn{1}{l|}{(0) Bad (Default); (1) Good (Paid)} &
  \multicolumn{1}{l|}{Label} \\ \hline
\multirow{2}{*}{Binary} &
  \multicolumn{1}{l|}{New Home} &
  \multicolumn{1}{l|}{Newly built dwelling} &
  \multicolumn{1}{l|}{(0) Old Home; (1) New Home} &
  \multicolumn{1}{l|}{NewHome} \\ \cline{2-5} 
 &
  \multicolumn{1}{l|}{First Time Buyer} &
  \multicolumn{1}{l|}{Never purchased property before} &
  \multicolumn{1}{l|}{(0) New Owner; (1) Previous Owner} &
  \multicolumn{1}{l|}{FTB} \\ \hline
\multirow{10}{*}{Continuous} &
  \multicolumn{1}{l|}{Age} &
  \multicolumn{1}{l|}{Age of the borrower} &
  \multicolumn{1}{l|}{18 - 55} &
  \multicolumn{1}{l|}{Age} \\ \cline{2-5} 
 &
  \multicolumn{1}{l|}{Income} &
  \multicolumn{1}{l|}{Total income of the borrower} &
  \multicolumn{1}{l|}{0 - inf} &
  \multicolumn{1}{l|}{Income} \\ \cline{2-5} 
 &
  \multicolumn{1}{l|}{Expenses-to-Income} &
  \multicolumn{1}{l|}{Ratio of borrower-expenditure-to-income} &
  \multicolumn{1}{l|}{0 - 1} &
  \multicolumn{1}{l|}{Exp:Inc} \\ \cline{2-5} 
 &
  \multicolumn{1}{l|}{Loan Value} &
  \multicolumn{1}{l|}{Amount advanced to the borrower} &
  \multicolumn{1}{l|}{0 - inf} &
  \multicolumn{1}{l|}{LoanValue} \\ \cline{2-5} 
 &
  \multicolumn{1}{l|}{Loan-to-Value} &
  \multicolumn{1}{l|}{Loan to value ratio} &
  \multicolumn{1}{l|}{1 - inf} &
  \multicolumn{1}{l|}{LTV} \\ \cline{2-5} 
 &
  \multicolumn{1}{l|}{Loan Term} &
  \multicolumn{1}{l|}{Length of the loan in years} &
  \multicolumn{1}{l|}{20 - 40} &
  \multicolumn{1}{l|}{LoanTerm} \\ \cline{2-5} 
 &
  \multicolumn{1}{l|}{Loan Rate} &
  \multicolumn{1}{l|}{Interest rate paid on the loan} &
  \multicolumn{1}{l|}{0 - 1} &
  \multicolumn{1}{l|}{Interest} \\ \cline{2-5} 
 &
  \multicolumn{1}{l|}{House Value} &
  \multicolumn{1}{l|}{Market value of the property} &
  \multicolumn{1}{l|}{1 - inf} &
  \multicolumn{1}{l|}{HouseVal} \\ \cline{2-5} 
 &
  \multicolumn{1}{l|}{MRTI} &
  \multicolumn{1}{l|}{Ratio of mortgage-repayments-to-income} &
  \multicolumn{1}{l|}{0 - 1} &
  \multicolumn{1}{l|}{MRTI} \\ \cline{2-5} 
 &
  \multicolumn{1}{l|}{Score} &
  \multicolumn{1}{l|}{Estimated credit risk score} &
  \multicolumn{1}{l|}{0 - inf} &
  \multicolumn{1}{l|}{Score} \\ \hline
\end{tabular}
\caption{Residential Mortgage Application Credit Scoring dataset description}
\label{tab:my-table}
\end{table}
\end{landscape}

\section{Conclusion} The objective of this short paper is to introduce a set of synthetic datasets to the community working on bias in ML. The data relates to mortgage applications and the model used to synthesise the data is based on a significant body of research on factors influencing mortgage approvals \cite{kennedy2011framework}. Altogether the datasets contain 37,607 examples described by 17 features. We focus on \emph{Age} as the sensitive feature and code is provided to convert this to a category. The threshold for this conversion can be adjusted to control the level of imbalance in the dataset. The level of imbalance in the outcome (Good/Bad) can also be controlled. We also provide exemplar datasets with 25\% class imbalance and 30\% feature imbalance. 

\bibliographystyle{plain}  
\bibliography{syn-data}

\begin{thebibliography}{1}

\bibitem{DBLP:journals/corr/abs-2104-14014}
William Blanzeisky and P{\'{a}}draig Cunningham.
\newblock Algorithmic factors influencing bias in machine learning.
\newblock {\em CoRR}, abs/2104.14014, 2021.

\bibitem{blanzeisky2021using}
William Blanzeisky and Pádraig Cunningham.
\newblock Using pareto simulated annealing to address algorithmic bias in
  machine learning, 2021.

\bibitem{caton2020fairness}
Simon Caton and Christian Haas.
\newblock Fairness in machine learning: A survey.
\newblock {\em arXiv preprint arXiv:2010.04053}, 2020.

\bibitem{Cunningham_2021}
Pádraig Cunningham and Sarah~Jane Delany.
\newblock Underestimation bias and underfitting in machine learning.
\newblock {\em Lecture Notes in Computer Science}, page 20–31, 2021.

\bibitem{feldman2015certifying}
Michael Feldman, Sorelle~A Friedler, John Moeller, Carlos Scheidegger, and
  Suresh Venkatasubramanian.
\newblock Certifying and removing disparate impact.
\newblock In {\em proceedings of the 21th ACM SIGKDD international conference
  on knowledge discovery and data mining}, pages 259--268, 2015.

\bibitem{holstein2019improving}
Kenneth Holstein, Jennifer Wortman~Vaughan, Hal Daum{\'e}~III, Miro Dudik, and
  Hanna Wallach.
\newblock Improving fairness in machine learning systems: What do industry
  practitioners need?
\newblock In {\em Proceedings of the 2019 CHI conference on human factors in
  computing systems}, pages 1--16, 2019.

\bibitem{kennedy2011framework}
Kenneth Kennedy, Sarah~Jane Delany, and Brian Mac~Namee.
\newblock A framework for generating data to simulate application scoring.
\newblock In {\em Credit Scoring and Credit Control XII, Conference
  Proceedings}. Credit Research Centre, Business School, University of
  Edinburgh, 2011.

\end{thebibliography}

\end{document}